\PassOptionsToPackage{hyphens}{url}
\RequirePackage[svgnames,table]{xcolor}
\documentclass[11pt,letterpaper,logo]{yalearxiv}

\usepackage[numbers,sort&compress]{natbib}
\usepackage{multirow,array,makecell,mathtools}
\usepackage{flafter,float}
\usepackage{tikz}
\usepackage{algorithm}
\usepackage{algpseudocode}
\allowdisplaybreaks[1]
\newsavebox{\tablebox}

\newcommand{\tablecheck}{\begin{tikzpicture}[x=1pt,y=1pt,baseline=-0.4pt]\draw[line width=0.5pt,line cap=round,line join=round] (0.3,2.1)--(1.8,0.6)--(5.2,4.5);\end{tikzpicture}}
\algrenewcommand{\algorithmicrequire}{\textbf{Input:}}
\algrenewcommand{\algorithmicensure}{\textbf{Output:}}
\algrenewcommand{\alglinenumber}[1]{\scriptsize #1:}

\title{Corroborative Learning: Mutual Enhancement through Task-Response Reconstruction}
\runningtitle{Corroborative Learning}
\author{Zexing Zhang, Jichao Li\textsuperscript{\dag}, Tianyang Lei,\\
\textbf{Yiyin Zhang}, and \textbf{Kewei Yang}\\
College of Systems Engineering, National University of Defense Technology\\
Changsha 410073, China}
\hypersetup{colorlinks=true,linkcolor=blue!50!black,citecolor=blue!50!black,urlcolor=blue!50!black,
  pdftitle={Corroborative Learning: Mutual Enhancement through Task-Response Reconstruction},
  pdfauthor={Zexing Zhang, Jichao Li, Tianyang Lei, Yiyin Zhang, Kewei Yang},
  pdfkeywords={corroborative learning, model collaboration, knowledge transfer, complementarity, independent inference}}

\begin{document}
\begin{abstract}
In pattern analysis, models trained independently for the same task often develop complementary capabilities through differences in architecture, data, and optimization. Exploiting complementarity while retaining independent deployment is central to collaborative learning. Existing research expands transferable knowledge through soft labels, intermediate representations, and sample relations, and improves compatibility through projections, cross-layer alignment, adaptive teaching, and bidirectional interaction. Most methods establish transfer by reducing differences in predictions, relations, or representations between target and collaborator, although these differences also underlie complementarity. This creates a structural trade-off between reducing differences for knowledge exchange and preserving them for complementarity. We therefore propose Corroborative Learning (CL). Each model forms task-driven responses, with cross-group support from its collaborator. Source-excluded reconstruction of response increments from collaborator representations separates fitting and evaluation across disjoint groups and selectively constrains directions by cross-group reconstructability. The process requires no direct output or feature matching or additional inference dependency. Furthermore, with linear reconstruction and fixed collaborator representations, the loss equals a positive semidefinite directional regularizer, constraining less reconstructable directions more strongly. Signal and noise conditions characterize when directional shrinkage reduces response-estimation error. Across six language tasks, 24 models, and 36 pairs, CL improves over capacity-matched independent adaptation on every task, with positive gains across most directed model combinations. Pairwise comparisons show broad advantages over transfer and mutual-learning methods and different gains from complementarity across collaboration rules. Low-label language and vision results support cross-domain applicability. Task-relevant differences can support mutual enhancement without eliminating disagreement.
\end{abstract}
\maketitle

\section{Introduction}\label{sec:1}
\begin{figure}[htb]
\centering
\includegraphics[width=\linewidth]{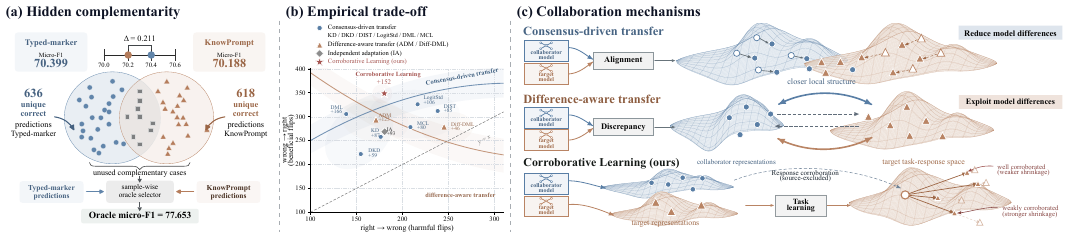}
\caption{Model complementarity and collaboration mechanisms. (a) Typed-marker and KnowPrompt differ by only 0.211 percentage points in micro-F1, yet have 636 and 618 unique correct predictions. A post hoc oracle gains 7.254 percentage points over the better model. (b) For KnowPrompt, corrections versus new errors under collaboration. The diagonal marks equal counts, and annotations show their difference. (c) Consensus alignment reduces model differences, difference-aware transfer exploits them, and source-excluded response corroboration constrains target changes.}\label{fig:complementarity_mechanisms}
\end{figure}

Models of different architectures and scales are often trained independently for the same task. Enabling them to improve one another while remaining separately deployable would extend the value of these existing models. Similar overall performance, however, does not imply the same capabilities. Differences in architecture, training data, augmentation, and optimization produce distinct representations and error patterns. A large-scale study of pretrained models finds that complementary knowledge is widespread and can even come from a model with lower overall performance \cite{ref1}. Model merging and reuse further show that differences among independently trained models can yield gains beyond selecting the strongest individual model \cite{ref2}. In cross-domain relation extraction, Typed-marker \cite{ref47} uses explicit entity markers, whereas KnowPrompt \cite{ref46} formulates relation prediction as prompt learning. On the same benchmark \cite{ref40}, their micro-F1 scores differ by only 0.211 percentage points, yet 636 and 618 examples are correctly predicted by only one of the two models, respectively. A post hoc oracle selecting predictions using ground-truth labels would improve micro-F1 by 7.254 percentage points over the better individual model, as shown in Fig.~\ref{fig:complementarity_mechanisms}(a). This raises a general question: can already trained models acquire complementary capabilities from each other while retaining their own deployment paths?

Knowledge-transfer research addresses this goal by enriching transferable information and improving compatibility between models. One line of work extends soft labels \cite{ref4} to intermediate features \cite{ref5}, sample relations \cite{ref7,ref8}, and contrastive representations \cite{ref9}. Knowledge decoupling and scale calibration further refine the information available for transfer \cite{ref10,ref11,ref12}. Our earlier consensus subspace distillation organizes transferable knowledge in a shared subspace \cite{ref62}. Recent large-scale evaluations also emphasize the information supplied to the student \cite{ref60}. Another improves transfer through cross-layer matching \cite{ref13}, cross-architecture projections \cite{ref16}, student feedback and adaptive teaching \cite{ref17,ref18,ref35}, and online mutual learning \cite{ref20,ref23}. Despite differences in knowledge and interaction, most of these methods establish transfer by reducing discrepancies between the predictions, relations, or representations of the target and collaborating models.

Alignment helps establish such a connection, but it also changes the differences on which complementarity depends. Model differences are both a source of mismatch in knowledge transfer and a source of information unavailable to an individual model. Online distillation can rapidly homogenize peers \cite{ref24}. IE-KD, Diff-DML, and ADM accordingly preserve exploratory features, predictive diversity, or a balance between consensus and divergence \cite{ref27,ref28,ref29}. As illustrated by the first two mechanisms in Fig.~\ref{fig:complementarity_mechanisms}(c), consensus-driven methods reduce model discrepancies through alignment, while difference-aware methods explicitly use disagreements. These findings reveal a structural trade-off: \textbf{reducing differences facilitates knowledge exchange, but continued convergence can weaken the differences that make collaboration useful.}

This trade-off is also visible in the resulting predictions. Figure~\ref{fig:complementarity_mechanisms}(b) compares methods by the errors corrected and the new errors introduced when KnowPrompt is the target. Consensus-driven approaches often correct more errors while also changing previously correct predictions, whereas difference-aware approaches exhibit a different balance. Model differences provide opportunities for exchange, but their value depends on which changes the target ultimately retains.

Existing methods primarily characterize \textbf{what knowledge a collaborator can provide}, whereas independently deployed performance depends on \textbf{how that knowledge changes the target's task function}. The same soft labels, relational structure, or projected representations can be matched through different function changes. These changes may fit the current supervision equally well but have different consequences on unseen data. Transferable knowledge therefore does not uniquely determine which target responses should be retained. How can a collaborator help identify which target changes remain useful on unseen data?

We propose \textbf{Corroborative Learning (CL)}. Each model first learns changes to its task scores from its own supervision. Its collaborator's representations are then used to reconstruct these changes. To test whether this relationship extends beyond the examples used to fit it, we partition the training data into non-overlapping groups. For each group, we fit the relationship on the remaining groups and evaluate reconstruction on that group. The resulting errors selectively constrain task-score changes, as illustrated in Fig.~\ref{fig:complementarity_mechanisms}(c). We call this source-excluded response reconstruction. With fixed collaborator representations, less reconstructable response directions receive stronger constraints. The two models are optimized jointly without directly matching outputs or intermediate features, and each remains independently deployable after training.

Our contributions are as follows:
\begin{itemize}
\setlength{\itemsep}{0pt}\setlength{\parsep}{0pt}
\widowpenalty=10000
\item We introduce \textbf{Corroborative Learning}, a perspective on model collaboration that addresses the trade-off between alignment and complementarity. It shifts the focus from which knowledge a peer supplies to which changes in the target model benefit from collaboration.
\item We develop \textbf{source-excluded response reconstruction}, a reciprocal learning mechanism that uses cross-group reconstructability to regulate each model's task-response increments. It supports different architectures, representation dimensions, and learning paradigms, retaining independent inference without directly matching outputs or intermediate features.
\item We characterize the \textbf{directional effect of corroboration}. With linear reconstruction and fixed peer representations, the loss is a positive semidefinite quadratic form. The analysis gives signal and noise conditions for directional shrinkage to reduce response-estimation error.
\item We evaluate CL on six language tasks with 24 independently trained models and 36 pairs, and on visual classification. Controlled analyses link gains to complementarity, peer information, and reconstruction behavior across supervision levels and model relationships.
\end{itemize}
\par

Section~\ref{sec:2} reviews related work on transferable knowledge, knowledge-transfer mechanisms, and diversity preservation. Section~\ref{sec:3} presents the CL method and its directional analysis. Section~\ref{sec:4} reports experiments on performance, complementarity, ablation, mechanisms, sensitivity, efficiency, and generalization to visual classification. Section~\ref{sec:5} concludes the paper.

\section{Related Work}\label{sec:2}
Model collaboration draws on trained-model reuse, representations of transferable knowledge, heterogeneous knowledge transfer, and mutual learning. These directions respectively model existing capabilities, source-side knowledge, cross-model compatibility, and relationships among peers. Table~\ref{tab:paradigms} summarizes their principal modeling dimensions. Existing work addresses model reuse, heterogeneity, reciprocal learning, and diversity preservation. CL focuses on the task-function changes induced in each target, since the same external knowledge can produce different responses in independently deployed models.

\subsection{Reuse and Integration of Trained Models}
Trained-model reuse primarily involves parameter merging and knowledge integration. Parameter merging operates directly on trained weights. Model Soups averages independently fine-tuned models \cite{ref2}. Recent work combines merging with layer-wise distillation \cite{ref65}, guides it through parameter extrapolation \cite{ref66}, and studies how gains scale with model size and expert count \cite{ref67}. Knowledge integration relaxes strict parameter compatibility. Pretrained-model analysis identifies task information absent from individual models \cite{ref1}, while Cooperative Knowledge Distillation organizes knowledge exchange according to model strengths \cite{ref37}. DUNE integrates heterogeneous teachers with different objectives and training data into a unified encoder \cite{ref68}. These approaches establish the reuse value of independently trained models at both parameter and functional levels. CL focuses on each target's task-score changes and evaluates them through cross-group reconstruction from peer representations.

\begin{table}[htb]
\centering
\caption{Settings and Modeling Focus of Collaboration Paradigms}
\label{tab:paradigms}
\begingroup
\fontsize{9}{10.8}\selectfont
\setlength{\tabcolsep}{0.7pt}
\renewcommand{\arraystretch}{1.28}
\begin{tabularx}{\linewidth}{@{}>{\raggedright\arraybackslash}X*{6}{>{\centering\arraybackslash}m{0.106\linewidth}}@{}}
\toprule
\rule{0pt}{11pt}Paradigm & {\fontsize{8}{9.5}\selectfont\makecell[c]{Indep.\\models}} & {\fontsize{8}{9.5}\selectfont\makecell[c]{Source\\info.}} & {\fontsize{8}{9.5}\selectfont\makecell[c]{Hetero.\\models}} & {\fontsize{8}{9.5}\selectfont\makecell[c]{Mutual\\learning}} & {\fontsize{8}{9.5}\selectfont\makecell[c]{Model\\diversity}} & {\fontsize{8}{9.5}\selectfont\makecell[c]{Response\\change}} \\
\midrule
Model merging \cite{ref2} & \tablecheck & \tablecheck &  &  & \tablecheck &  \\
Knowledge integration \cite{ref37,ref68} & \tablecheck & \tablecheck & \tablecheck & \tablecheck &  &  \\
Output distillation \cite{ref4,ref10} &  & \tablecheck & \tablecheck &  &  &  \\
Feature distillation \cite{ref5} &  & \tablecheck & \tablecheck &  &  &  \\
Relation distillation \cite{ref7,ref8,ref31} &  & \tablecheck & \tablecheck &  &  &  \\
Heterogeneous distillation \cite{ref17,ref18} &  & \tablecheck & \tablecheck &  &  &  \\
Mutual learning \cite{ref20,ref23} &  & \tablecheck & \tablecheck & \tablecheck &  &  \\
Difference-aware learning \cite{ref27,ref28,ref29,ref63} &  & \tablecheck & \tablecheck & \tablecheck & \tablecheck &  \\
Corroborative Learning & \tablecheck & \tablecheck & \tablecheck & \tablecheck & \tablecheck & \tablecheck \\
\bottomrule
\end{tabularx}
\endgroup
\par\smallskip\noindent\parbox{\linewidth}{\footnotesize Checkmarks summarize the settings and emphases of the cited formulations. Blank cells do not imply incompatibility. Indep. and hetero. denote independently trained and heterogeneous models. Source info. denotes transferable knowledge.}
\end{table}

\subsection{Transferable Knowledge Representations}
Distillation transfers responses, features, or relations \cite{ref4,ref5,ref7}. Response-based methods extend soft labels \cite{ref4} through DKD's target/non-target decoupling \cite{ref10}, DIST's prediction relations \cite{ref11}, and Logit Standardization's scale calibration \cite{ref12}. Recent logit-based methods model finer class relations \cite{ref69} and covariance-normalized output directions \cite{ref70}. Feature-based methods transfer hidden representations and contrastive information beyond final predictions \cite{ref5,ref9,ref62}. Relation-based methods use sample distances, angles, similarities, or correlations to reduce dependence on compatible absolute feature spaces \cite{ref7,ref8,ref31}. These methods enrich transferable information and reduce reliance on direct prediction matching. Their constraints are nevertheless defined by source-side knowledge. The same prediction, feature, or relation constraint can induce different task-function changes in different targets. CL defines the collaboration constraint on the target's task-response increment and evaluates its reconstruction from peer representations.

\subsection{Knowledge Transfer Mechanisms}
Architectural and capacity differences motivate representation alignment and adaptive teaching. Alignment makes heterogeneous representations more comparable. SemCKD learns cross-layer correspondences from semantic correlations \cite{ref13}, and Cross-Architecture KD studies knowledge mapping across model families \cite{ref16}. Receptive-field alignment further addresses spatial mismatch between teacher and student features \cite{ref61}. Progressive residual approximation addresses teacher-student capacity gaps \cite{ref73}. Adaptive teaching instead adjusts the teaching signal to the student. PAT adapts heterogeneous teacher features through student feedback \cite{ref17}, and DIST+ and adaptive-teacher distillation further reduce mismatches between teacher knowledge and student needs \cite{ref18,ref35}. These strategies improve representation compatibility and teaching suitability, broadening the scope of heterogeneous transfer. Targets still differ in parameterization and inductive bias, so a compatible transfer constraint can be realized by different responses with different effects on unseen examples. CL models these responses directly and uses cross-group reconstruction to constrain changes without aligning the two feature spaces.

\subsection{Consensus and Differences in Mutual Learning}
Mutual learning combines consensus-driven interaction with diversity preservation. DML jointly improves peers through mutual prediction matching \cite{ref20}. MCL extends online interaction through mutual contrastive relations \cite{ref23}. Recent work also learns knowledge-transfer directions between graph networks \cite{ref63}.  Sustained consensus can also homogenize models. OKDDip observes rapid convergence under simple online aggregation \cite{ref24}, and subsequent methods maintain diversity through knowledge inheritance and exploration and differentiated mutual learning \cite{ref27,ref28}. ADM jointly uses feature consensus and divergence \cite{ref29}. These studies establish that model differences can have collaborative value. Diversity is generally regulated through relationships among predictions or representations, but its magnitude alone does not distinguish useful complementarity from model-specific information or prediction errors. Multi-view co-regularization jointly learns predictors by encouraging agreement on unlabeled examples \cite{ref74}. CL uses peer representations to reconstruct each target's task-score increments, with each group excluded when fitting the reconstruction relation.

\section{Corroborative Learning}\label{sec:3}
\subsection{Framework and Problem Formulation}\label{sec:3_1}
\begin{figure}[htb]
\centering
\includegraphics[width=\linewidth]{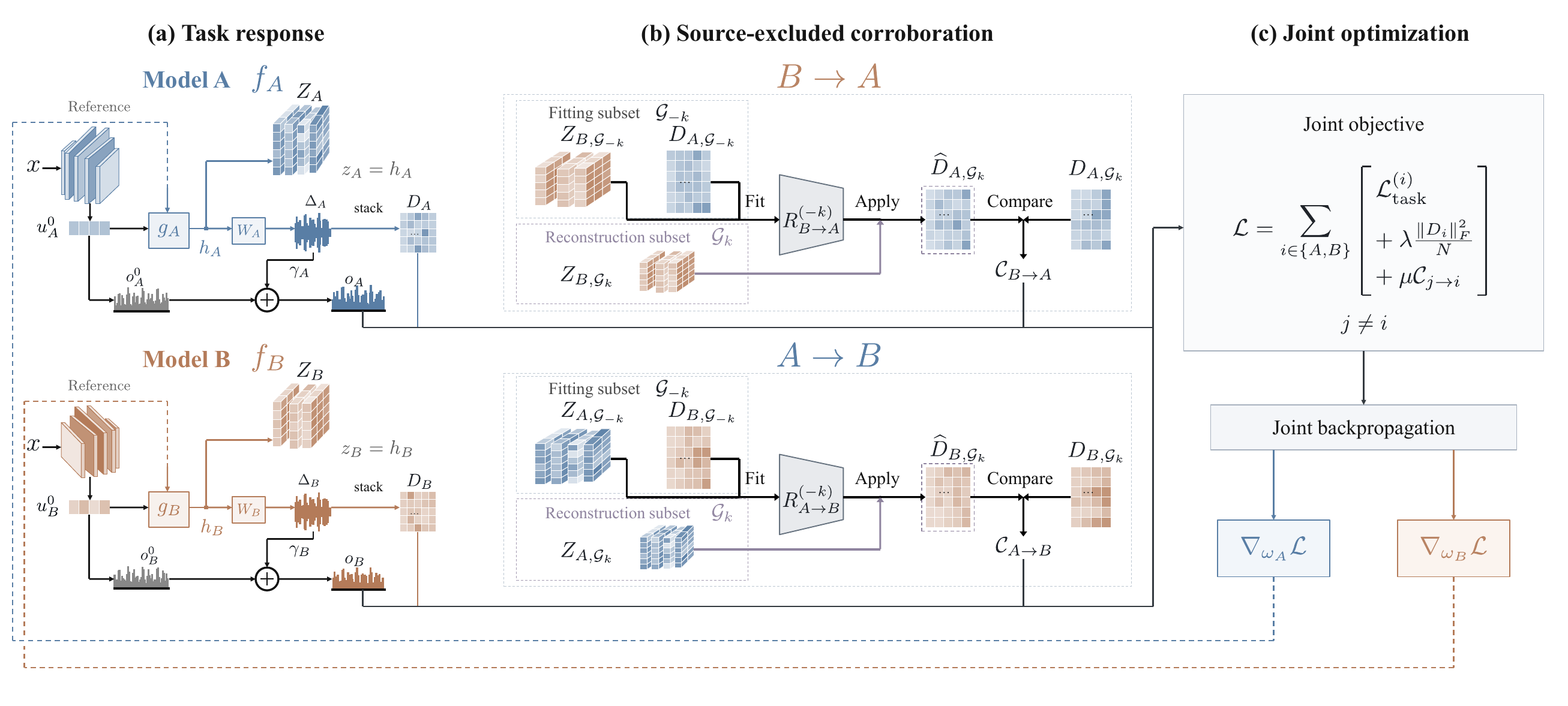}
\caption{Corroborative Learning framework. (a) The two models form task-response increments $D_i$ and representations $Z_i$. (b) For each excluded group $\mathcal G_k$, a relation fitted on the remaining groups reconstructs its responses. Errors form the two corroboration losses $\mathcal C_{j\rightarrow i}$. (c) Task losses and bidirectional corroboration jointly update both models. Repeated symbols denote group slices.}\label{fig2_method_overview}
\end{figure}

CL jointly adapts two independently trained models while preserving their separate prediction paths. Each model learns task-score changes, and the other supplies representations for reconstructing them. Figure~\ref{fig2_method_overview} outlines response construction, reconstruction across data groups, and joint optimization.

The models process corresponding training inputs. We partition these inputs into disjoint groups for reconstruction fitting and evaluation, while supervised adaptation uses all groups.

Let the supervised dataset be

\[
\mathcal D=\{(x_n,y_n)\}_{n=1}^{N},
\tag{1}\label{eq:cl_1}
\]
where \(x_n\) is an input and \(y_n\) its supervision. Corroborative Learning also specifies a partition

\[
\mathcal P=\{\mathcal G_1,\ldots,\mathcal G_K\},
\qquad
\bigcup_{k=1}^{K}\mathcal G_k=\{1,\ldots,N\},
\tag{2}\label{eq:cl_2}
\]
where \(2\le K\le N\), with nonempty, pairwise disjoint groups. For each \(\mathcal G_k\), reconstruction is fitted on the remaining examples \(\mathcal G_{-k}\) and evaluated on \(\mathcal G_k\). This defines source-excluded reconstruction. Section~\ref{sec:4_1} describes task-specific grouping.

Let \(f_1^0,f_2^0\) denote the independently trained reference models, whose architectures and sizes may differ. Model \(i\in\{1,2\}\) supplies fixed reference scores \(o_i^0(x)\in\mathbb R^{r_i}\), updated scores \(o_i(x)\), and collaborator representations \(z_i(x)\in\mathbb R^{p_i}\). These dimensions need not match across models. Direction \(j\!\rightarrow\!i\) uses model \(j\)'s representations to reconstruct model \(i\)'s score changes on corresponding examples.

The task-response branch in Fig.~\ref{fig2_method_overview}(a) uses a trainable adapter to produce \(o_i(x)\). A fixed scale \(\gamma_i>0\) normalizes the change as \(\Delta_i(x)=[o_i(x)-o_i^0(x)]/\gamma_i\). Stacking the increments and collaborator representations gives \(D_i\in\mathbb R^{N\times r_i}\) and \(Z_i\in\mathbb R^{N\times p_i}\). Section~\ref{sec:3_2} defines the adapter.

Reconstruction in Fig.~\ref{fig2_method_overview}(b) fits \(R_{j\rightarrow i}^{(-k)}\) from \(Z_{j,\mathcal G_{-k}}\) to \(D_{i,\mathcal G_{-k}}\), then reconstructs \(\widehat D_{i,\mathcal G_k}\) from \(Z_{j,\mathcal G_k}\). Summing the group errors gives the corroboration loss \(\mathcal C_{j\rightarrow i}\). Exchanging roles gives \(\mathcal C_{i\rightarrow j}\). Section~\ref{sec:3_3} defines the relation and loss.

Joint optimization in Fig.~\ref{fig2_method_overview}(c) combines each model's task loss, an amplitude penalty on score changes, and both corroboration losses into \(\mathcal L\). It updates both adapters, yielding final model parameters \(\theta_1^\star,\theta_2^\star\), as detailed in Section~\ref{sec:3_4}. After training, model \(i\) predicts
\[
\hat y_i(x)=\operatorname{Decode}_i(o_i(x)),
\tag{3}\label{eq:cl_3}
\]
where \(\operatorname{Decode}_i\) is the task decoder. Each model uses its own scores \(o_i(x)\) and prediction path at inference.

\subsection{Task-Response Modeling}\label{sec:3_2}
\begin{figure}[htb]
\centering
\includegraphics[width=0.58\linewidth]{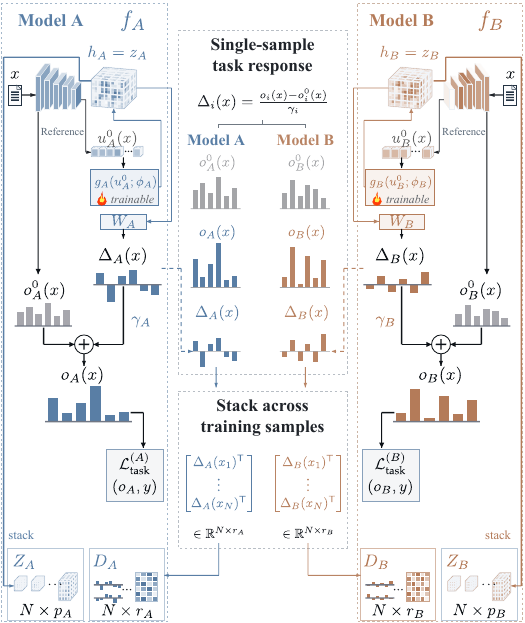}
\caption{Task-response construction. Each adapter produces a normalized increment $\Delta_i(x)$ that augments fixed reference scores $o_i^0(x)$. Stacked increments and hidden representations yield $D_i$ and $Z_i$ for reconstruction.}\label{fig3_task_response_modeling}
\end{figure}

As shown in Fig.~\ref{fig3_task_response_modeling}, each model starts from fixed task scores and learns an additive response through its own adapter. We model changes in task scores because they directly affect predictions. Parameter and feature changes depend on model-specific spaces, while a score increment measures how collaboration changes each model's task function.

For model \(i\) and a fixed response scale \(\gamma_i>0\) derived from its reference scores, define the scale-normalized task-response increment

\[
\Delta_i(x)=\frac{o_i(x)-o_i^0(x)}{\gamma_i}.
\tag{4}\label{eq:cl_4}
\]
The increment remains in model \(i\)'s task space. Its dimension, scale, and implementation may differ from those of its peer. CL does not require equal increments across models.

Let \(u_i(x)\) be a feature near the task readout and \(a_i(u_i(x);\omega_i)\in\mathbb R^{r_i}\) a differentiable adapter. Parameters \(\omega_i\) are trainable during collaboration. \(\theta_i\) denotes all model parameters, including \(\omega_i\). Then

\[
\Delta_i(x)=a_i(u_i(x);\omega_i),
\tag{5}\label{eq:cl_5}
\]
\[
o_i(x)=o_i^0(x)+\gamma_i\Delta_i(x),
\tag{6}\label{eq:cl_6}
\]
We use a lightweight residual adapter with a 64-dimensional GELU hidden layer and fixed reference features \(u_i(x)=u_i^0(x)\) in \eqref{eq:cl_5}:
\[
\begin{aligned}
h_i(x)&=g_i(u_i^0(x);\phi_i), & \Delta_i(x)&=W_i h_i(x),\\
z_i(x)&=h_i(x), & \omega_i&=(\phi_i,W_i).
\end{aligned}
\]
where \(h_i,z_i\in\mathbb R^{p_i}\) and \(W_i\in\mathbb R^{r_i\times p_i}\). The trainable hidden state \(h_i\) supplies both the response mapping and corroborating representation. It is recomputed at each update. Task supervision and both reconstruction terms update \(\omega_i\), while reference scores remain fixed. Only \(W_i\) is zero-initialized, giving \(\Delta_i(x)=0\) and \(o_i(x)=o_i^0(x)\) initially.
Score changes are \(\gamma_i\Delta_i\), while amplitude and corroboration penalties use \(\Delta_i\). Other differentiable adapters may be used if they realize \eqref{eq:cl_4}.

Model \(i\) retains its original task supervision

\[
\mathcal L_{\mathrm{task}}^{(i)}
=
\frac{1}{N}\sum_{n=1}^{N}\ell_i(o_i(x_n),y_n),
\tag{7}\label{eq:cl_7}
\]
where \(\ell_i\) is the supervised task loss. It directs the response toward the model's own task objective.

The lower panels of Fig.~\ref{fig3_task_response_modeling} show how sample-wise responses and representations form the matrices \(D_i\) and \(Z_i\) used for reconstruction.

\subsection{Source-Excluded Response Corroboration}\label{sec:3_3}
\begin{figure}[htb]
\centering
\includegraphics[width=\linewidth]{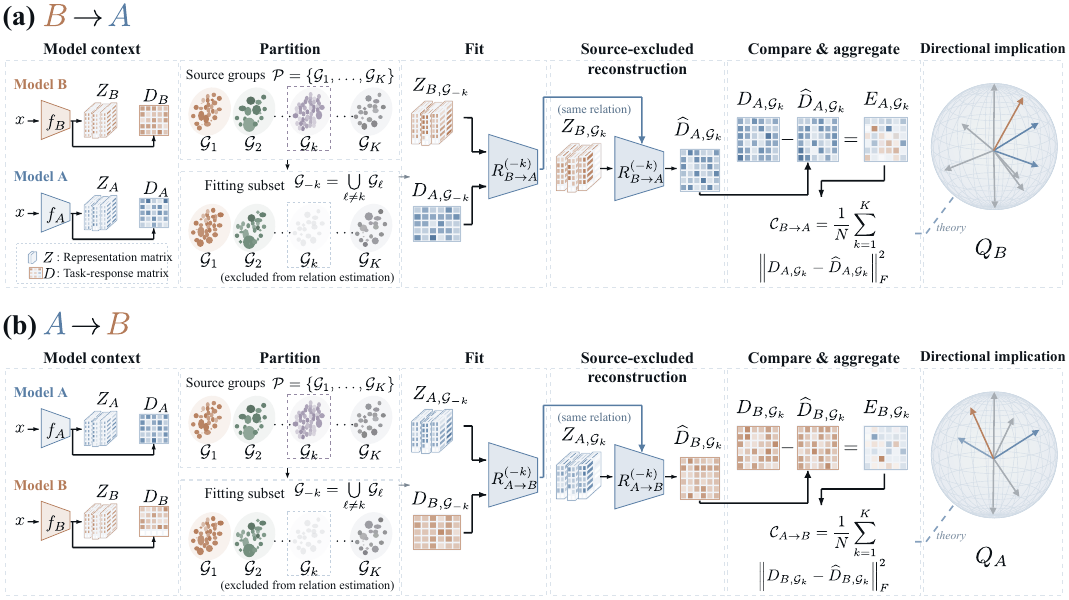}
\caption{Source-excluded corroboration in both directions. (a) The relation $R_{B\rightarrow A}^{(-k)}$, fitted outside $\mathcal G_k$, reconstructs the withheld target responses to form $\mathcal C_{B\rightarrow A}$. (b) Exchanging model roles forms $\mathcal C_{A\rightarrow B}$. Residual patterns determine the directional operator $Q_j$.}\label{fig4_source_excluded_corroboration}
\end{figure}

Figure~\ref{fig4_source_excluded_corroboration} expands the source-excluded fit and evaluation for both directions. A reconstructor fitted and evaluated on the same examples can exploit incidental associations. We therefore exclude each target group from fitting and evaluate reconstruction on that group, measuring whether the relationship extends across groups.

Stack model \(i\)'s responses over all training examples as

\[
D_i=
[\Delta_i(x_1),\ldots,\Delta_i(x_N)]^\top
\in\mathbb R^{N\times r_i},
\tag{8}\label{eq:cl_8}
\]
and model \(j\)'s corroborating representations as

\[
Z_j=
[z_j(x_1),\ldots,z_j(x_N)]^\top
\in\mathbb R^{N\times p_j}.
\tag{9}\label{eq:cl_9}
\]
Rows of \(D_i\) and \(Z_j\) correspond to the same examples (Fig.~\ref{fig3_task_response_modeling}).

For group \(\mathcal G_k\), define its complement

\[
\mathcal G_{-k}
=
\{1,\ldots,N\}\setminus\mathcal G_k.
\tag{10}\label{eq:cl_10}
\]
The relationship for direction \(j\!\rightarrow\!i\) is estimated using only \(\mathcal G_{-k}\):

\[
R_{j\rightarrow i}^{(-k)}
=
\mathcal A
\left(
Z_{j,\mathcal G_{-k}},
D_{i,\mathcal G_{-k}}
\right),
\tag{11}\label{eq:cl_11}
\]
where \(\mathcal A\) is differentiable. The fitted reconstructor is applied to the excluded group without refitting:

\[
\widehat D_{i,\mathcal G_k}
=
R_{j\rightarrow i}^{(-k)}
\left(
Z_{j,\mathcal G_k}
\right).
\tag{12}\label{eq:cl_12}
\]
Target responses in \(\mathcal G_k\) do not enter the fit. Backpropagation retains the fitting graph, allowing gradients to reach fitting responses through \eqref{eq:cl_11} or its linear instance \eqref{eq:cl_ridge}. Low reconstruction error on \(\mathcal G_k\) indicates a relationship that extends beyond the fitting groups, while task supervision guides the predictive utility of the resulting responses.

We use ridge regression for an explicit, differentiable linear reconstructor:

\[
B_{j\rightarrow i}^{(-k)}
=
\left(
\frac{
Z_{j,\mathcal G_{-k}}^\top Z_{j,\mathcal G_{-k}}
}{
|\mathcal G_{-k}|
}
+
\rho I
\right)^{-1}
\frac{
Z_{j,\mathcal G_{-k}}^\top D_{i,\mathcal G_{-k}}
}{
|\mathcal G_{-k}|
},
\tag{13}\label{eq:cl_ridge}
\]
where \(\rho>0\) is the ridge coefficient and \(I\) is the identity matrix. Reconstruction is

\[
\widehat D_{i,\mathcal G_k}
=
Z_{j,\mathcal G_k}B_{j\rightarrow i}^{(-k)}.
\tag{14}\label{eq:cl_14}
\]
Aggregating \(K\) groups gives the corroboration loss for \(j\!\rightarrow\!i\):

\[
\mathcal C_{j\rightarrow i}
=
\frac{1}{N}\sum_{k=1}^{K}
\left\|
D_{i,\mathcal G_k}
-
\widehat D_{i,\mathcal G_k}
\right\|_F^2.
\tag{15}\label{eq:cl_15}
\]
This loss measures cross-group reconstruction of target responses. In Fig.~\ref{fig4_source_excluded_corroboration}, the upper and lower paths produce \(\mathcal C_{B\rightarrow A}\) and \(\mathcal C_{A\rightarrow B}\), respectively, from the exchanged roles of target responses and peer representations.

\subsection{Reciprocal Optimization and Directional Interpretation}\label{sec:3_4}
\begingroup
\setlength{\abovedisplayskip}{6pt plus 1pt minus 1pt}
\setlength{\belowdisplayskip}{6pt plus 1pt minus 1pt}
\setlength{\abovedisplayshortskip}{3pt plus 2pt minus 2pt}
\setlength{\belowdisplayshortskip}{6pt plus 1pt minus 1pt}

Figure~\ref{fig5_reciprocal_directional_effect}(a) combines the task, amplitude, and peer-corroboration losses in both model branches. Their sum defines the joint objective

\[
\mathcal L
=
\sum_{i=1}^{2}
\left[
\mathcal L_{\mathrm{task}}^{(i)}
+
\lambda\frac{\|D_i\|_F^2}{N}
+
\mu\mathcal C_{3-i\rightarrow i}
\right],
\tag{16}\label{eq:cl_16}
\]
Weights \(\lambda\ge0\) and \(\mu\ge0\) control response amplitude and corroboration strength. Both directions use the same pre-update state and are differentiated jointly. Setting \(\mu=0\) recovers independent adaptation with the same response parameterization and amplitude term.

\begin{figure}[htb]
\centering
\includegraphics[width=\linewidth]{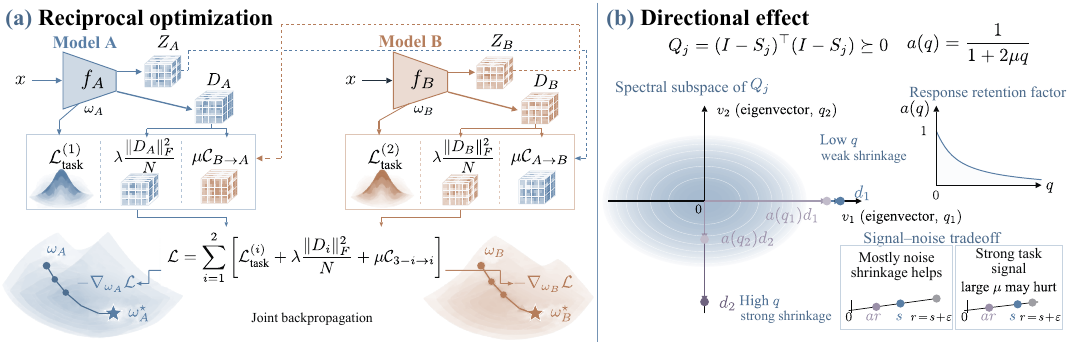}
\caption{Reciprocal optimization and directional interpretation. (a) Both adapters are jointly updated by task, amplitude, and peer-corroboration losses. (b) For fixed peer representations, the eigenvalues $q$ of $Q_j\succeq0$ govern directional penalties. The local benefit of the resulting shrinkage depends on signal and noise.}\label{fig5_reciprocal_directional_effect}
\end{figure}

After training, all \(R_{j\rightarrow i}^{(-k)}\), \(B_{j\rightarrow i}^{(-k)}\), and cross-model connections are discarded. The resulting models with parameters \(\theta_1^\star,\theta_2^\star\) predict independently.

For the directional analysis, fix the current peer representations, partition \(\mathcal P\), and coefficient \(\rho\). The reconstruction in \eqref{eq:cl_ridge}--\eqref{eq:cl_14} is then linear in \(D_i\), so an operator \(S_j\) satisfies

\[
\widehat D_i=S_jD_i.
\tag{17}\label{eq:cl_17}
\]
Consequently,

\[
\mathcal C_{j\rightarrow i}
=
\frac{1}{N}
\left\|(I-S_j)D_i\right\|_F^2
=
\frac{1}{N}
\operatorname{tr}(D_i^\top Q_jD_i),
\tag{18}\label{eq:cl_18}
\]
where

\[
Q_j=(I-S_j)^\top(I-S_j)\succeq0.
\tag{19}\label{eq:cl_19}
\]
Matrix \(Q_j\in\mathbb R^{N\times N}\) acts along the sample dimension. Its eigenvectors describe response patterns across examples. Figure~\ref{fig5_reciprocal_directional_effect}(b) illustrates how the eigenvalues govern directional penalties, as formalized below.

Proposition 1 (Directional Effect of Corroboration). With linear reconstruction, fixed peer representations, and a fixed partition, the corroboration loss is a positive semidefinite quadratic form. Response directions with larger eigenvalues of \(Q_j\) have greater cross-group reconstruction error per unit response energy and receive stronger constraints.

For fixed collaborator representations, the gradient with respect to \(D_i\) is

\[
\nabla_{D_i}\mathcal C_{j\rightarrow i}
=
\frac{2}{N}Q_jD_i.
\tag{20}\label{eq:cl_20}
\]
During training, peer representations change with the adapter parameters. The fixed-representation analysis characterizes the directional effect at a given state.

Consider an eigen-direction of \(Q_j\). Without the corroboration term, write its response as

\[
r=s+\varepsilon,
\]
where \(s\) is the task-relevant signal and \(\varepsilon\) is a zero-mean perturbation with variance \(v\). For the corresponding eigenvalue \(q\), a unit-curvature local quadratic surrogate centered at this reference response is
\[
\min_t\ \frac{1}{2N}(t-r)^2+\frac{\mu q}{N}t^2.
\]
Its minimizer is \(t^\star=ar\), with response retention factor

\[
a=\frac{1}{1+2\mu q}.
\tag{21}\label{eq:cl_21}
\]
The expected estimation error relative to \(s\) is

\[
\mathbb E(ar-s)^2
=
(1-a)^2s^2+a^2v.
\tag{22}\label{eq:cl_22}
\]
When \(\mu q>0\), so that \(0<a<1\), the error in \eqref{eq:cl_22} is strictly smaller than the unconstrained error \(v\) if and only if

\[
v>
\frac{1-a}{1+a}s^2.
\tag{23}\label{eq:cl_23}
\]
If \(\mu q=0\), then \(a=1\) and the direction is unchanged. Otherwise, shrinkage reduces error exactly when \eqref{eq:cl_23} holds. Stronger corroboration can suppress noise but also useful model-specific signal, motivating the directional and weight analyses in Section~\ref{sec:4_6}.

\endgroup
\setcounter{equation}{23}
\section{Experiments and Analysis}\label{sec:4}

The experiments address seven connected questions about performance, complementarity, ablation, mechanisms, sensitivity, efficiency, and generalization. We first establish the gains, then examine their sources and behavior, and finally assess their cost and scope.

\begin{enumerate}
\item \textbf{Q1 Performance.} Does CL improve independent model performance? Section~\ref{sec:4_2} compares six tasks and all pairs.
\item \textbf{Q2 Complementarity.} How do collaboration rules use existing complementarity? Section~\ref{sec:4_3} compares the same initial pairs.
\item \textbf{Q3 Ablation.} Which components contribute to the gains? Section~\ref{sec:4_4} isolates collaborator information, source exclusion, and reciprocal optimization.
\item \textbf{Q4 Mechanisms.} How do response changes relate to task utility and directional regularization? Sections~\ref{sec:4_5} and~\ref{sec:4_6} examine samples and spectral directions.
\item \textbf{Q5 Sensitivity.} How do supervision, grouping, and model relationships affect the gains? Section~\ref{sec:4_7} varies these conditions.
\item \textbf{Q6 Efficiency.} What training and inference costs accompany the gains? Section~\ref{sec:4_8} evaluates both.
\item \textbf{Q7 Generalization.} Do the gains extend from language tasks to visual classification? Section~\ref{sec:4_9} tests this extension.
\end{enumerate}

\subsection{Experimental Setup}\label{sec:4_1}
\subsubsection*{Datasets and Models}
We evaluate Re-TACRED \cite{ref39} and CrossRE \cite{ref40} for sentence-level and cross-domain relation extraction, MNLI \cite{ref41} for natural language inference, Re-DocRED \cite{ref42} for document-level relation extraction, and ACE05 \cite{ref43} and WikiEvents \cite{ref44} for event argument extraction. These NLP benchmarks vary in context granularity, output structure, and source organization. Together with the diverse model pool, they test whether a common collaboration rule can improve existing predictors that represent and solve a task differently.

\begin{figure}[htb]
\centering
\includegraphics[width=\linewidth]{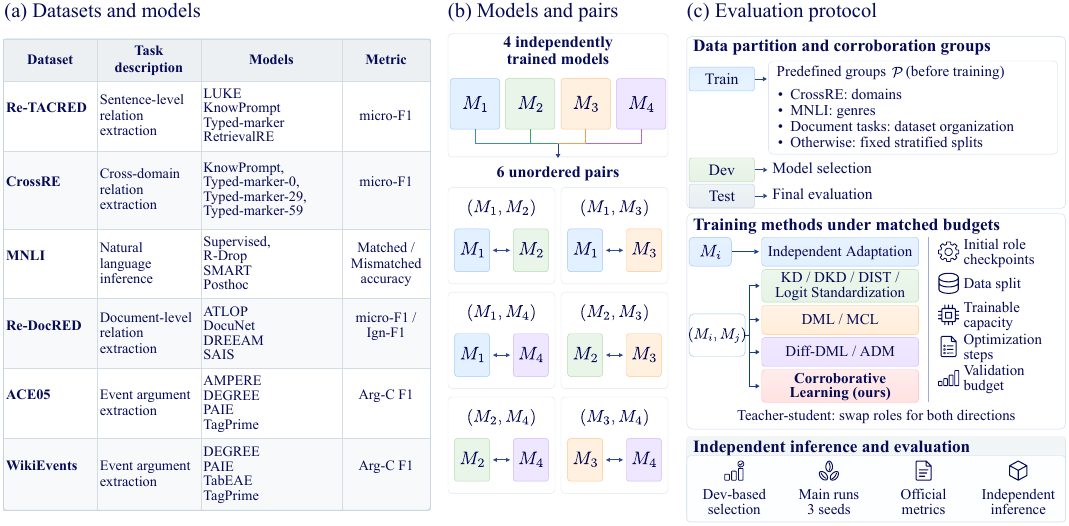}
\caption{Experimental setup and evaluation protocol. (a) Datasets, models, and metrics. Dataset sources are cited in the text. (b) Model pairing. (c) Matched evaluation protocol. Three-seed aggregation applies to the main experiments.}\label{fig6_experimental_setup}
\end{figure}

Each task has four independently trained models, covering different architectures and discriminative, prompt-based, retrieval-augmented, or generative methods. On CrossRE, Typed-marker runs trained with different random seeds allow complementarity within a fixed model formulation to be evaluated alongside differences between marker-based and prompt-based models. Figure~\ref{fig6_experimental_setup}(a) lists the models and metrics.

Four models yield six unordered pairs per task. Each pair is evaluated in both directions: 36 pairs and 72 directed results per random seed. Each target has multiple peers, allowing comparison across model relationships.

\subsubsection*{Corroboration Groups}

Groups use available source metadata: CrossRE domains \cite{ref40}, MNLI genres \cite{ref41}, and native document organization for document-level tasks. Where reliable source identifiers are unavailable, predefined stratified partitions provide an alternative. Groups are fixed before training. Alternatives are evaluated in Section~\ref{sec:4_7}.

\subsubsection*{Baselines and Training Protocol}

Independent Adaptation (IA) uses the same response parameterization and amplitude penalty as CL, without peer information. Transfer baselines are KD \cite{ref4}, DKD \cite{ref10}, DIST \cite{ref11}, and Logit Standardization \cite{ref12}. Mutual-learning baselines are DML \cite{ref20} and MCL \cite{ref23}. Diff-DML \cite{ref28} preserves predictive diversity, and ADM \cite{ref29} combines feature consensus and divergence.

To isolate the effect of each collaboration rule, methods share initial models, data partitions, trainable capacity, update steps, and numbers of validation settings. Test metrics use official benchmark scripts. One-way transfer and Diff-DML are run separately with roles exchanged. Main results average validation-selected runs over three random seeds. The \(\mu\) and \(K\) sweeps each use a fixed random seed: \(\mu\) uses the same-configuration \(\mu=0\) adaptation as reference, while \(K\) uses the seed's validation-selected IA state. Cross-task summaries average within-task gains, not raw metrics. Within each sweep, other training settings are held fixed. Baseline objectives are evaluated within this shared adapter protocol.

\subsection{Performance Across Models}\label{sec:4_2}

For Q1, the initial Checkpoint measures the change from each trained model, while capacity-matched Independent Adaptation (IA) separates improvements from further task training and those associated with collaboration. Table~\ref{tab:full_comparison_means} compares methods under the same protocol. Its main values average the 12 directed combinations in each task, and its pairwise-win column summarizes comparisons at the matched directed-combination level.

\begin{table}[htb]
\centering
\caption{Performance across six tasks. Main values are means over three seeds and 12 directed combinations per task. Superscripts report gains over the target Checkpoint for a fixed representative direction. Subscripts report the across-seed standard deviation of the direction-averaged score. Gains and standard deviations are in percentage points. Bold and underlined main values indicate the best and second-best results.}
\label{tab:full_comparison_means}
\begingroup
\definecolor{PFHeader}{HTML}{EAF1FF}
\definecolor{PFBaseline}{HTML}{F6F7F9}
\definecolor{PFStripe}{HTML}{F6F7F9}
\definecolor{PFWhite}{HTML}{FFFFFF}
\definecolor{PFEmphasis}{HTML}{EAF3FF}
\definecolor{PFRule}{HTML}{353C46}
\definecolor{PFPositive}{HTML}{0066CC}
\definecolor{PFNegative}{HTML}{B95426}
\definecolor{PFNeutral}{HTML}{868C92}
\definecolor{PFWinHeader}{HTML}{EAF1FF}
\definecolor{PFGroup1}{HTML}{FFF0E6}
\definecolor{PFGroup2}{HTML}{EAF2F8}
\definecolor{PFGroup3}{HTML}{EEF5FF}
\definecolor{PFGroup4}{HTML}{F0EDFF}
\definecolor{PFGroup5}{HTML}{F3EEFF}
\definecolor{PFGroup6}{HTML}{FFF3E8}
\fontsize{8.6}{11}\selectfont
\renewcommand{\arraystretch}{1.08}
\setlength{\tabcolsep}{0pt}
\setlength{\aboverulesep}{0pt}\setlength{\belowrulesep}{0pt}
\arrayrulecolor{PFRule}
\newlength{\PFMethodWidth}\newlength{\PFMetricWidth}\newlength{\PFWinWidth}
\setlength{\PFMethodWidth}{0.12\linewidth}
\setlength{\PFWinWidth}{0.085\linewidth}
\setlength{\PFMetricWidth}{\dimexpr(\linewidth-\PFMethodWidth-\PFWinWidth)/8\relax}
\newcommand{\PFscore}[3]{\makebox[24pt][r]{{\fontsize{8}{10}\selectfont #1}}\hspace{0.3pt}\makebox[17.5pt][l]{$^{\text{\fontsize{5.7}{6.5}\selectfont #2}}_{\text{\fontsize{5.7}{6.5}\selectfont #3}}$}}
\begin{tabular}{@{}>{\centering\arraybackslash}m{\PFMethodWidth}*{8}{>{\centering\arraybackslash}m{\PFMetricWidth}}>{\centering\arraybackslash}m{\PFWinWidth}@{}}
\toprule[0.9pt]
\rowcolor{PFHeader}  & \multicolumn{1}{>{\centering\arraybackslash}m{\dimexpr 1\PFMetricWidth\relax}}{\cellcolor{PFGroup1}\rule[-5pt]{0pt}{17pt}{\fontsize{8}{10}\selectfont\textbf{CrossRE}}} & \multicolumn{1}{>{\centering\arraybackslash}m{\dimexpr 1\PFMetricWidth\relax}}{\cellcolor{PFGroup2}\rule[-5pt]{0pt}{17pt}{\fontsize{8}{10}\selectfont\textbf{Re-TACRED}}} & \multicolumn{2}{>{\centering\arraybackslash}m{\dimexpr 2\PFMetricWidth\relax}}{\cellcolor{PFGroup3}\rule[-5pt]{0pt}{17pt}{\fontsize{8}{10}\selectfont\textbf{Re-DocRED}}} & \multicolumn{2}{>{\centering\arraybackslash}m{\dimexpr 2\PFMetricWidth\relax}}{\cellcolor{PFGroup4}\rule[-5pt]{0pt}{17pt}{\fontsize{8}{10}\selectfont\textbf{MNLI}}} & \multicolumn{1}{>{\centering\arraybackslash}m{\dimexpr 1\PFMetricWidth\relax}}{\cellcolor{PFGroup5}\rule[-5pt]{0pt}{17pt}{\fontsize{8}{10}\selectfont\textbf{WikiEvents}}} & \multicolumn{1}{>{\centering\arraybackslash}m{\dimexpr 1\PFMetricWidth\relax}}{\cellcolor{PFGroup6}\rule[-5pt]{0pt}{17pt}{\fontsize{8}{10}\selectfont\textbf{ACE05}}} & \cellcolor{PFWinHeader} \\
\rowcolor{PFHeader} \multirow{-2}{\PFMethodWidth}{\centering\textbf{Method}} & \cellcolor{PFGroup1}{\fontsize{7.2}{9}\selectfont\shortstack{micro-F1\\(\%) $\uparrow$}} & \cellcolor{PFGroup2}{\fontsize{7.2}{9}\selectfont\shortstack{micro-F1\\(\%) $\uparrow$}} & \cellcolor{PFGroup3}{\fontsize{7.2}{9}\selectfont\shortstack{micro-F1\\(\%) $\uparrow$}} & \cellcolor{PFGroup3}{\fontsize{7.2}{9}\selectfont\shortstack{Ign-F1\\(\%) $\uparrow$}} & \cellcolor{PFGroup4}{\fontsize{7.2}{9}\selectfont\shortstack{Matched\\acc. (\%) $\uparrow$}} & \cellcolor{PFGroup4}{\fontsize{7.2}{9}\selectfont\shortstack{Mismatched\\acc. (\%) $\uparrow$}} & \cellcolor{PFGroup5}{\fontsize{7.2}{9}\selectfont\shortstack{Arg-C F1\\(\%) $\uparrow$}} & \cellcolor{PFGroup6}{\fontsize{7.2}{9}\selectfont\shortstack{Arg-C F1\\(\%) $\uparrow$}} & \cellcolor{PFWinHeader}\multirow{-2}{\PFWinWidth}{\centering{\fontsize{7.7}{9}\selectfont\shortstack{\textbf{Pairwise}\\\textbf{Win}}}} \\
\midrule[0.65pt]
\rowcolor{PFBaseline} \rule[-2pt]{0pt}{15pt}Checkpoint & \rule[-2pt]{0pt}{15pt}\PFscore{70.549}{}{0.147} & \rule[-2pt]{0pt}{15pt}\PFscore{91.036}{}{0.061} & \rule[-2pt]{0pt}{15pt}\PFscore{78.172}{}{0.074} & \rule[-2pt]{0pt}{15pt}\PFscore{77.404}{}{0.083} & \rule[-2pt]{0pt}{15pt}\PFscore{89.137}{}{0.037} & \rule[-2pt]{0pt}{15pt}\PFscore{89.051}{}{0.049} & \rule[-2pt]{0pt}{15pt}\PFscore{59.323}{}{0.204} & \rule[-2pt]{0pt}{15pt}\PFscore{71.201}{}{0.143} & \rule[-2pt]{0pt}{15pt}318/960 \\
\midrule
\rowcolor{PFWhite} \rule[-2pt]{0pt}{15pt}IA & \rule[-2pt]{0pt}{15pt}\PFscore{71.565}{\textcolor{PFPositive}{+0.528}}{0.189} & \rule[-2pt]{0pt}{15pt}\PFscore{89.286}{\textcolor{PFPositive}{+0.164}}{0.173} & \rule[-2pt]{0pt}{15pt}\PFscore{78.171}{\textcolor{PFPositive}{+0.153}}{0.104} & \rule[-2pt]{0pt}{15pt}\PFscore{77.393}{\textcolor{PFPositive}{+0.127}}{0.112} & \rule[-2pt]{0pt}{15pt}\PFscore{88.991}{\textcolor{PFNegative}{-0.125}}{0.063} & \rule[-2pt]{0pt}{15pt}\PFscore{88.870}{\textcolor{PFNegative}{-0.108}}{0.072} & \rule[-2pt]{0pt}{15pt}\PFscore{59.761}{\textcolor{PFPositive}{+2.579}}{0.253} & \rule[-2pt]{0pt}{15pt}\PFscore{71.108}{\textcolor{PFPositive}{+0.060}}{0.181} & \rule[-2pt]{0pt}{15pt}186/960 \\
\midrule
\rowcolor{PFStripe} \rule[-2pt]{0pt}{15pt}KD & \rule[-2pt]{0pt}{15pt}\PFscore{71.082}{\textcolor{PFPositive}{+0.563}}{0.162} & \rule[-2pt]{0pt}{15pt}\PFscore{91.126}{\textcolor{PFPositive}{+0.236}}{0.077} & \rule[-2pt]{0pt}{15pt}\PFscore{\textbf{78.499}}{\textcolor{PFNeutral}{+0.000}}{0.095} & \rule[-2pt]{0pt}{15pt}\PFscore{\textbf{77.772}}{\textcolor{PFNeutral}{+0.000}}{0.106} & \rule[-2pt]{0pt}{15pt}\PFscore{\underline{89.185}}{\textcolor{PFPositive}{+0.021}}{0.056} & \rule[-2pt]{0pt}{15pt}\PFscore{\underline{89.090}}{\textcolor{PFPositive}{+0.070}}{0.062} & \rule[-2pt]{0pt}{15pt}\PFscore{59.911}{\textcolor{PFPositive}{+2.593}}{0.225} & \rule[-2pt]{0pt}{15pt}\PFscore{71.186}{\textcolor{PFPositive}{+0.099}}{0.157} & \rule[-2pt]{0pt}{15pt}551/960 \\
\rowcolor{PFWhite} \rule[-2pt]{0pt}{15pt}LogitStd & \rule[-2pt]{0pt}{15pt}\PFscore{71.559}{\textcolor{PFPositive}{+1.326}}{0.171} & \rule[-2pt]{0pt}{15pt}\PFscore{91.089}{\textcolor{PFPositive}{+0.193}}{0.082} & \rule[-2pt]{0pt}{15pt}\PFscore{78.274}{\textcolor{PFPositive}{+0.194}}{0.091} & \rule[-2pt]{0pt}{15pt}\PFscore{77.464}{\textcolor{PFPositive}{+0.135}}{0.103} & \rule[-2pt]{0pt}{15pt}\PFscore{89.108}{\textcolor{PFPositive}{+0.027}}{0.053} & \rule[-2pt]{0pt}{15pt}\PFscore{88.993}{\textcolor{PFPositive}{+0.048}}{0.061} & \rule[-2pt]{0pt}{15pt}\PFscore{59.971}{\textcolor{PFPositive}{+2.638}}{0.218} & \rule[-2pt]{0pt}{15pt}\PFscore{71.231}{\textcolor{PFPositive}{+0.257}}{0.158} & \rule[-2pt]{0pt}{15pt}464/960 \\
\rowcolor{PFStripe} \rule[-2pt]{0pt}{15pt}DIST & \rule[-2pt]{0pt}{15pt}\PFscore{71.337}{\textcolor{PFPositive}{+0.869}}{0.165} & \rule[-2pt]{0pt}{15pt}\PFscore{\underline{91.146}}{\textcolor{PFPositive}{+0.235}}{0.089} & \rule[-2pt]{0pt}{15pt}\PFscore{78.268}{\textcolor{PFPositive}{+0.230}}{0.107} & \rule[-2pt]{0pt}{15pt}\PFscore{77.488}{\textcolor{PFPositive}{+0.206}}{0.117} & \rule[-2pt]{0pt}{15pt}\PFscore{89.052}{\textcolor{PFNegative}{-0.031}}{0.061} & \rule[-2pt]{0pt}{15pt}\PFscore{88.933}{\textcolor{PFPositive}{+0.010}}{0.068} & \rule[-2pt]{0pt}{15pt}\PFscore{59.937}{\textcolor{PFPositive}{+2.704}}{0.238} & \rule[-2pt]{0pt}{15pt}\PFscore{71.174}{\textcolor{PFPositive}{+0.140}}{0.176} & \rule[-2pt]{0pt}{15pt}396/960 \\
\rowcolor{PFWhite} \rule[-2pt]{0pt}{15pt}DKD & \rule[-2pt]{0pt}{15pt}\PFscore{71.550}{\textcolor{PFPositive}{+0.880}}{0.173} & \rule[-2pt]{0pt}{15pt}\PFscore{91.112}{\textcolor{PFPositive}{+0.210}}{0.085} & \rule[-2pt]{0pt}{15pt}\PFscore{78.283}{\textcolor{PFPositive}{+0.240}}{0.099} & \rule[-2pt]{0pt}{15pt}\PFscore{77.502}{\textcolor{PFPositive}{+0.216}}{0.112} & \rule[-2pt]{0pt}{15pt}\PFscore{88.999}{\textcolor{PFNegative}{-0.091}}{0.057} & \rule[-2pt]{0pt}{15pt}\PFscore{88.869}{\textcolor{PFNegative}{-0.090}}{0.066} & \rule[-2pt]{0pt}{15pt}\PFscore{59.883}{\textcolor{PFPositive}{+2.662}}{0.231} & \rule[-2pt]{0pt}{15pt}\PFscore{71.183}{\textcolor{PFNegative}{-0.038}}{0.168} & \rule[-2pt]{0pt}{15pt}388/960 \\
\midrule
\rowcolor{PFStripe} \rule[-2pt]{0pt}{15pt}DML & \rule[-2pt]{0pt}{15pt}\PFscore{\textbf{71.949}}{\textcolor{PFPositive}{\underline{+1.590}}}{0.154} & \rule[-2pt]{0pt}{15pt}\PFscore{91.134}{\textcolor{PFPositive}{\underline{+0.295}}}{0.079} & \rule[-2pt]{0pt}{15pt}\PFscore{78.325}{\textcolor{PFPositive}{\underline{+0.342}}}{0.087} & \rule[-2pt]{0pt}{15pt}\PFscore{77.506}{\textcolor{PFPositive}{\underline{+0.283}}}{0.096} & \rule[-2pt]{0pt}{15pt}\PFscore{89.180}{\textcolor{PFPositive}{\underline{+0.122}}}{0.048} & \rule[-2pt]{0pt}{15pt}\PFscore{89.070}{\textcolor{PFPositive}{\underline{+0.152}}}{0.055} & \rule[-2pt]{0pt}{15pt}\PFscore{59.959}{\textcolor{PFPositive}{\underline{+2.785}}}{0.207} & \rule[-2pt]{0pt}{15pt}\PFscore{\underline{71.268}}{\textcolor{PFPositive}{\underline{+0.281}}}{0.149} & \rule[-2pt]{0pt}{15pt}\underline{689/960} \\
\rowcolor{PFWhite} \rule[-2pt]{0pt}{15pt}MCL & \rule[-2pt]{0pt}{15pt}\PFscore{71.487}{\textcolor{PFPositive}{+1.467}}{0.168} & \rule[-2pt]{0pt}{15pt}\PFscore{91.142}{\textcolor{PFPositive}{+0.237}}{0.083} & \rule[-2pt]{0pt}{15pt}\PFscore{78.340}{\textcolor{PFPositive}{+0.239}}{0.092} & \rule[-2pt]{0pt}{15pt}\PFscore{77.519}{\textcolor{PFPositive}{+0.184}}{0.105} & \rule[-2pt]{0pt}{15pt}\PFscore{89.130}{\textcolor{PFPositive}{+0.048}}{0.055} & \rule[-2pt]{0pt}{15pt}\PFscore{89.010}{\textcolor{PFPositive}{+0.068}}{0.064} & \rule[-2pt]{0pt}{15pt}\PFscore{59.885}{\textcolor{PFPositive}{+2.075}}{0.216} & \rule[-2pt]{0pt}{15pt}\PFscore{71.190}{\textcolor{PFPositive}{+0.234}}{0.158} & \rule[-2pt]{0pt}{15pt}527/960 \\
\rowcolor{PFStripe} \rule[-2pt]{0pt}{15pt}ADM & \rule[-2pt]{0pt}{15pt}\PFscore{71.590}{\textcolor{PFPositive}{+1.279}}{0.181} & \rule[-2pt]{0pt}{15pt}\PFscore{91.072}{\textcolor{PFPositive}{+0.047}}{0.086} & \rule[-2pt]{0pt}{15pt}\PFscore{78.164}{\textcolor{PFNeutral}{+0.000}}{0.108} & \rule[-2pt]{0pt}{15pt}\PFscore{77.389}{\textcolor{PFNeutral}{+0.000}}{0.116} & \rule[-2pt]{0pt}{15pt}\PFscore{89.140}{\textcolor{PFNegative}{-0.010}}{0.059} & \rule[-2pt]{0pt}{15pt}\PFscore{89.011}{\textcolor{PFNeutral}{+0.000}}{0.065} & \rule[-2pt]{0pt}{15pt}\PFscore{\underline{60.031}}{\textcolor{PFPositive}{+2.618}}{0.236} & \rule[-2pt]{0pt}{15pt}\PFscore{71.231}{\textcolor{PFNegative}{-0.129}}{0.172} & \rule[-2pt]{0pt}{15pt}397/960 \\
\rowcolor{PFWhite} \rule[-2pt]{0pt}{15pt}Diff-DML & \rule[-2pt]{0pt}{15pt}\PFscore{71.230}{\textcolor{PFPositive}{+1.209}}{0.177} & \rule[-2pt]{0pt}{15pt}\PFscore{91.104}{\textcolor{PFPositive}{+0.156}}{0.092} & \rule[-2pt]{0pt}{15pt}\PFscore{78.164}{\textcolor{PFNeutral}{+0.000}}{0.110} & \rule[-2pt]{0pt}{15pt}\PFscore{77.395}{\textcolor{PFNeutral}{+0.000}}{0.122} & \rule[-2pt]{0pt}{15pt}\PFscore{89.114}{\textcolor{PFNeutral}{+0.000}}{0.065} & \rule[-2pt]{0pt}{15pt}\PFscore{88.985}{\textcolor{PFPositive}{+0.010}}{0.073} & \rule[-2pt]{0pt}{15pt}\PFscore{\textbf{60.034}}{\textcolor{PFPositive}{+2.618}}{0.244} & \rule[-2pt]{0pt}{15pt}\PFscore{71.160}{\textcolor{PFPositive}{+0.273}}{0.185} & \rule[-2pt]{0pt}{15pt}352/960 \\
\midrule
\rowcolor{PFEmphasis} \rule[-2pt]{0pt}{15pt}\textbf{CL (Ours)} & \rule[-2pt]{0pt}{15pt}\PFscore{\underline{71.847}}{\textcolor{PFPositive}{\textbf{+1.596}}}{0.142} & \rule[-2pt]{0pt}{15pt}\PFscore{\textbf{91.208}}{\textcolor{PFPositive}{\textbf{+0.305}}}{0.070} & \rule[-2pt]{0pt}{15pt}\PFscore{\underline{78.410}}{\textcolor{PFPositive}{\textbf{+0.343}}}{0.080} & \rule[-2pt]{0pt}{15pt}\PFscore{\underline{77.617}}{\textcolor{PFPositive}{\textbf{+0.310}}}{0.091} & \rule[-2pt]{0pt}{15pt}\PFscore{\textbf{89.245}}{\textcolor{PFPositive}{\textbf{+0.133}}}{0.043} & \rule[-2pt]{0pt}{15pt}\PFscore{\textbf{89.140}}{\textcolor{PFPositive}{\textbf{+0.168}}}{0.051} & \rule[-2pt]{0pt}{15pt}\PFscore{59.973}{\textcolor{PFPositive}{\textbf{+2.829}}}{0.195} & \rule[-2pt]{0pt}{15pt}\PFscore{\textbf{71.318}}{\textcolor{PFPositive}{\textbf{+0.308}}}{0.136} & \rule[-2pt]{0pt}{15pt}\textbf{827/960} \\
\bottomrule[0.9pt]
\end{tabular}
\arrayrulecolor{black}
\endgroup
\par\vspace{3pt}\noindent\parbox{\linewidth}{\fontsize{7.1}{9}\selectfont Pairwise Win: Results are first averaged over three random seeds for each directed combination and metric, then compared with the other ten methods using unrounded values. Each method has $8\times12\times10=960$ comparisons. Strictly larger values count as wins, and ties do not.}
\end{table}

CL exceeds Checkpoint on all eight metrics, with a mean gain of +0.43 percentage points after averaging metrics within each task and weighting the six tasks equally. The gains include +1.30 on CrossRE and +0.65 on WikiEvents. Across the methods in Table~\ref{tab:full_comparison_means}, CL ranks first on four metrics and second on three, spanning relation extraction, inference, and event argument extraction. On WikiEvents, it remains close to the leading alternatives while exceeding its own Checkpoint by +0.65 percentage points. Across all eight metrics, CL has the lowest across-seed standard deviation among the adaptation methods.

Relative to capacity-matched IA, CL improves all six tasks, with a mean gain of +0.52 percentage points. On Re-TACRED, CL exceeds IA by 1.922 points and the initial Checkpoint by 0.172 points. Looking beyond task averages, CL records 827 of 960 strict pairwise wins against the other ten methods, the highest aggregate in the table. DML records 689.

\begin{figure}[htb]
\centering

\includegraphics[width=\linewidth]{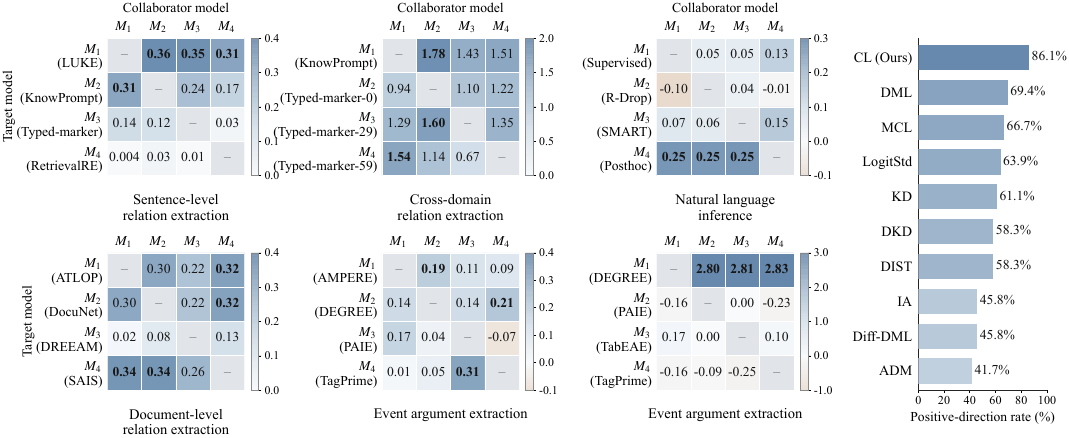}
\caption{Directed collaboration gains and positive-gain direction rates across six tasks. Read left to right and top to bottom, the heatmaps correspond to Re-TACRED, CrossRE, MNLI, Re-DocRED, ACE05, and WikiEvents. Heatmaps show CL's percentage-point changes relative to the target Checkpoint for 12 directed combinations per task. Rows are targets and columns are collaborators. MNLI uses the mean of matched and mismatched accuracy, Re-DocRED uses micro-F1, and the other tasks use the corresponding metric in Table~\ref{tab:full_comparison_means}. Each task has its own color scale. The right panel summarizes ten methods across all 72 directions. A gain is positive only when its unrounded value is strictly greater than zero.}\label{fig7_overall_effectiveness}
\end{figure}

Figure~\ref{fig7_overall_effectiveness} shows where these gains arise. Across 36 pairs, CL improves 62 of 72 directed results (86.1\%) over Checkpoint, compared with 69.4\% positive directions for DML. All twelve CrossRE directions improve, although their magnitudes differ. The heatmaps show that gains depend on both collaborator choice and collaboration direction. This variation motivates the next question: how much of a pair's available complementarity does each collaboration rule translate into a benefit?

\subsection{Complementarity and Collaboration Gains}\label{sec:4_3}

Overall gains do not show whether a method benefits from the differences that exist before collaboration. For Q2, we therefore compare methods on the same initial pairs, separating their shared complementary capacity from the gains produced by each collaboration rule.

The six initial CrossRE pairs have mean prediction disagreement of 18.53\%. 11.77\% of all test examples are correct for only one member. Gold-label oracle selection provides 5.52 percentage points of mean headroom over each pair's stronger member. Collaboration must realize this potential through target-model changes that can either correct errors or damage correct predictions.

\begin{figure}[htb]
\centering

\includegraphics[width=\linewidth]{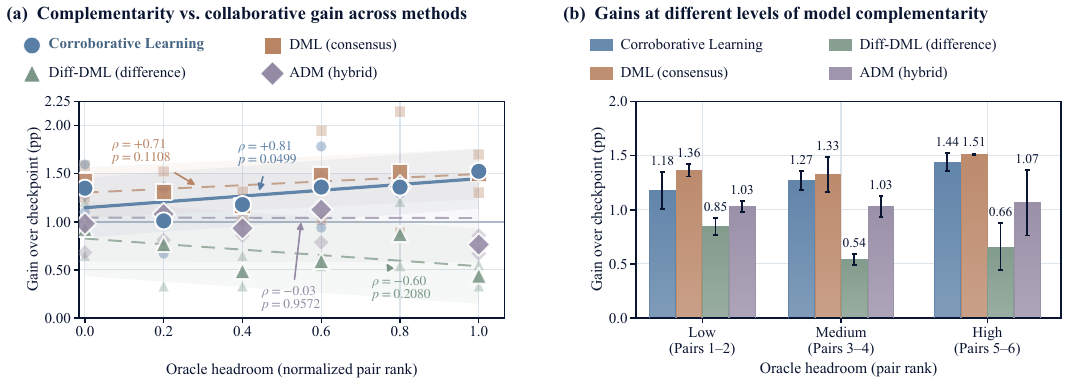}
\caption{Model complementarity and conversion into gains under different mechanisms.
(a) Complementary capacity and collaboration gains for six initial CrossRE pairs. Correlations use the mean gain over both directions of each pair. The horizontal axis is normalized oracle-headroom rank. Larger markers are pair means and lighter markers are directed results. Trend lines and shading denote linear fits and their 95\% confidence intervals.
(b) Mean gains at different complementarity levels. Error bars are standard errors over model pairs.}\label{fig8_complementarity_gain}
\end{figure}

For each directed result of method \(m\), define
\begin{equation}
	\Delta_m=M_m-M_{\mathrm{Checkpoint}},\label{eq:gain}
\end{equation}
where \(M_m\) and \(M_{\mathrm{Checkpoint}}\) are target performance after and before collaboration. Across the six CrossRE pairs in Fig.~\ref{fig8_complementarity_gain}(a), pair-level gains are positively associated with oracle headroom for CL (Spearman \(\rho=0.81\)) and DML (\(0.71\)), negatively associated for Diff-DML (\(-0.60\)), and weakly associated for ADM (\(-0.03\)). This descriptive comparison suggests that collaboration rules use the same available complementarity differently.

From low to high complementarity, CL gains +1.18, +1.27, and +1.44 percentage points, as shown in Fig.~\ref{fig8_complementarity_gain}(b). DML has larger absolute gains but a different group pattern. Diff-DML gains less in the high-complementarity group, while ADM is nearly stable. Absolute performance and sensitivity to complementarity capture different aspects of collaboration.

\begin{table}[b]
\centering
\caption{Ablation of Corroborating Information Relative to IA}\label{tab:attribution}
\begingroup
\fontsize{8.5}{10.2}\selectfont
\setlength{\tabcolsep}{1.25pt}
\renewcommand{\arraystretch}{1.22}
\begin{tabularx}{\linewidth}{@{}>{\raggedright\arraybackslash}X>{\centering\arraybackslash}p{0.112\linewidth}*{3}{>{\centering\arraybackslash}p{0.084\linewidth}}>{\raggedleft\arraybackslash}p{0.18\linewidth}>{\raggedleft\arraybackslash}p{0.125\linewidth}@{}}
\toprule
Variant & \makecell[c]{Relation\\source} & \makecell[c]{Held-\\out} & \makecell[c]{Peer\\upd.} & \makecell[c]{Two-\\way} & \makecell[r]{Gain\\(pp)} & \makecell[r]{Positive\\(\%)} \\
\midrule
Independent adaptation & -- & -- & -- & -- & 0.00 & -- \\
Self-reconstruction & Self & \tablecheck & -- & -- & \(\text{+0.14}_{\text{\fontsize{6.1}{7}\selectfont 0.082}}\) & 59.7 \\
Same-source recon. & Peer & -- & \tablecheck & \tablecheck & \(\text{+0.23}_{\text{\fontsize{6.1}{7}\selectfont 0.088}}\) & 66.7 \\
Shuffled pairs & Shuffled & \tablecheck & \tablecheck & \tablecheck & \(\text{+0.03}_{\text{\fontsize{6.1}{7}\selectfont 0.079}}\) & 52.8 \\
Frozen collaborator & Peer & \tablecheck & -- & -- & \(\text{\underline{+0.37}}_{\text{\fontsize{6.1}{7}\selectfont 0.091}}\) & \underline{76.4} \\
One-way corroboration & Peer & \tablecheck & \tablecheck & -- & \(\text{+0.34}_{\text{\fontsize{6.1}{7}\selectfont 0.097}}\) & 73.6 \\
\textbf{Complete method} & Peer & \tablecheck & \tablecheck & \tablecheck & \(\text{\textbf{+0.52}}_{\text{\fontsize{6.1}{7}\selectfont 0.086}}\) & \textbf{88.9} \\
\bottomrule
\end{tabularx}
\endgroup
\par\smallskip\noindent\parbox{\linewidth}{\fontsize{7.8}{9.2}\selectfont Relation source distinguishes target features (Self), correctly paired peer features (Peer), and shuffled peer correspondence (Shuffled). Check marks indicate excluded-group fitting, updates to the collaborator, and two active corroboration directions. Subscripts give the across-seed standard deviations of the six-task mean gains. Positive denotes the percentage of 72 directed results improving on IA. Bold and underlined values mark the best and second-best.}
\end{table}

These differences motivate attribution: Section~\ref{sec:4_4} tests whether the gains require genuine peer information, source exclusion, and reciprocal optimization.

\subsection{Ablation of Corroborating Information}\label{sec:4_4}

For Q3, we isolate the contributions of peer information, excluded-group reconstruction, and reciprocal updating. Table~\ref{tab:attribution} varies these components while holding the data, task supervision, trainable response capacity, update steps, and model selection fixed.

Self-reconstruction uses target features, shuffled pairing breaks correspondence, and Same-source reconstruction evaluates the relation on its fitting group. Frozen collaborator holds peer adaptation parameters fixed, while One-way corroboration retains one directed loss. All variants keep their respective task and amplitude losses. Self-reconstruction gains +0.14 even without peer information. Same-source reconstruction, which fits and evaluates within the same group, gains +0.23. Shuffling peer correspondence reduces the gain to +0.03. With matched peer information and excluded-group fitting, Frozen collaborator and One-way corroboration gain +0.37 and +0.34, respectively. The complete method combines these ingredients in reciprocal training, reaching +0.52 with 88.9\% positive directions.

Correct peer correspondence and source-excluded fitting each improve performance, supporting CL's use of cross-group reconstruction to guide target changes. Section~\ref{sec:4_5} next analyzes whether reconstruction quality identifies useful target changes at the sample level.

\subsection{Sample-Level Mechanism}\label{sec:4_5}

For Q4, we relate reconstruction quality to sample-level task utility, accounting for both corrected errors and changes to previously correct predictions.

\begin{figure}[htb]
\centering

\includegraphics[width=0.48\linewidth]{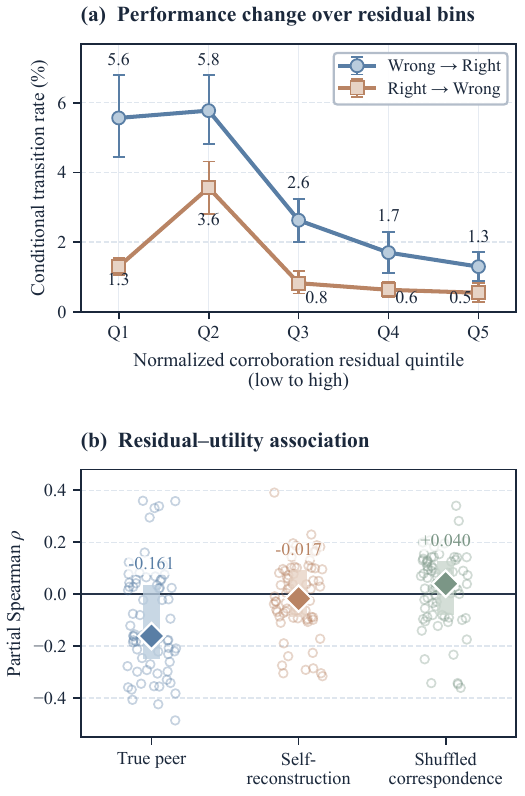}
\caption{Corroboration quality and task utility. (a) Conditional rates of beneficial wrong-to-right and harmful right-to-wrong changes across five normalized-residual quintiles. Denominators are initially incorrect and initially correct examples within each group, respectively. Error bars are 95\% bootstrap confidence intervals for the displayed aggregate conditional rates. (b) Partial Spearman correlations between residual and utility under true-peer corroboration, self-reconstruction, and shuffled correspondence. Open circles denote directed results, diamonds denote medians, and colored bars denote interquartile ranges.}\label{fig:corroboration_quality}
\end{figure}

For a sample $x_n$, define the normalized corroboration residual
\begin{equation}
	e_n =\label{eq:residual}
	\frac{\|\Delta_i(x_n)-\widehat{\Delta}_i(x_n)\|_2}
	{\|\Delta_i(x_n)\|_2+\epsilon},
\end{equation}
where $\epsilon$ ensures stability. We record gold-label margin changes for classification, decoder-consistent score changes for structured or generative tasks, and loss and prediction changes for all tasks. Test-set utility is the decrease in supervised loss:
\[
U_n=\ell_i(o_i^0(x_n),y_n)-\ell_i(o_i(x_n),y_n).
\]
Partial Spearman correlation is computed within each directed combination, controlling for initial loss and response amplitude \(\|\Delta_i(x_n)\|_2\). Correction and new-error rates are conditioned on initially incorrect and initially correct examples within each residual group, respectively.

Across five residual quintiles, correction rates fall from 5.6\% and 5.8\% in the first two to 1.3\% in the last. New-error rates peak at 3.6\% in the second and fall to 0.5\%, as shown in Fig.~\ref{fig:corroboration_quality}(a). Low residuals accompany more changes of both kinds, so residual magnitude alone does not determine utility.

The median partial correlation is $-0.161$ with genuine peer information, versus $-0.017$ for self-reconstruction and $+0.040$ for shuffled correspondence, as shown in Fig.~\ref{fig:corroboration_quality}(b). The stronger negative association suggests that genuine peer information helps distinguish more useful response changes.

The sample analysis connects residuals with outcomes. We next examine how corroboration acts across response directions, following Section~\ref{sec:3_4}.

\subsection{Directional Mechanism}\label{sec:4_6}

We examine the directional behavior described by Proposition~1 and the local risk condition in \eqref{eq:cl_23} using spectral directions of \(Q_j\) computed through implicit matrix-vector products and randomized eigendecomposition. Within eigenvalue quantiles, response-amplitude attenuation is
\[
\eta=1-\sqrt{\frac{E_{\mathrm{after}}}{E_{\mathrm{before}}}},
\qquad E_{\mathrm{before}}>0,
\]
where the energies are measured along the same direction before and after comparison. Re-TACRED contributes 12 directions for each of 12 directed combinations, totaling 144. In a fixed analysis basis, repeated-response means provide a signal reference and repeated-response variances estimate noise. Separate evaluation repetitions measure error relative to this reference. Both states use the same pre-corroboration signal energy for normalization.

\begin{figure}[htb]
\centering
\includegraphics[width=0.54\linewidth]{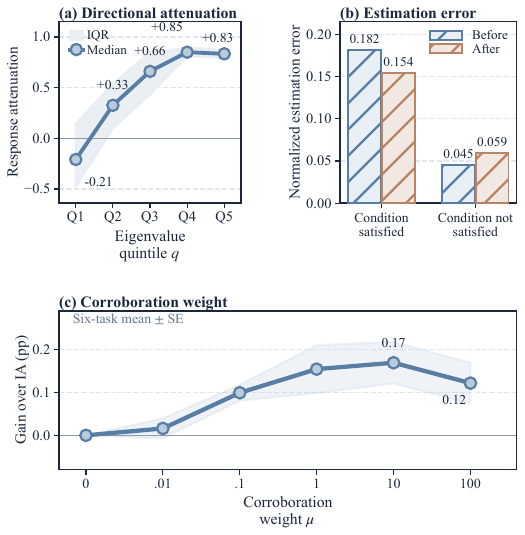}
\caption{Directional behavior and theoretical consistency.
(a) Response-amplitude attenuation in selected Re-TACRED spectral directions: 12 per directed combination and 144 in total. Curves and shading show medians and interquartile ranges within eigenvalue quantiles.
(b) Directional signal, noise, and shrinkage are estimated from experimental data. Directions are grouped by the condition in \eqref{eq:cl_23} to compare normalized estimation error before and after corroboration.
(c) Fixed-seed six-task mean gains relative to matched $\mu=0$ adaptation for different corroboration weights $\mu$. Shading denotes the standard error across tasks.}\label{fig:directional_behavior}
\end{figure}

Attenuation generally increases with eigenvalue \(q\), reaching approximately 0.85 and 0.83 in the highest two quantiles, as shown in Fig.~\ref{fig:directional_behavior}(a). Grouping by \eqref{eq:cl_23}, normalized error decreases from 0.182 to 0.154 when the condition holds and increases from 0.045 to 0.059 otherwise, as shown in Fig.~\ref{fig:directional_behavior}(b). These results are consistent with the local analysis: shrinkage reduces estimation error when noise suppression outweighs signal loss.

In the fixed-seed \(\mu\) sweep with a common training endpoint, the six-task mean gain relative to the matched \(\mu=0\) configuration peaks among the tested values at approximately +0.17 percentage points for \(\mu=10\), then falls to approximately +0.12 at \(\mu=100\), as shown in Fig.~\ref{fig:directional_behavior}(c). This nonmonotonic pattern is consistent with the signal and noise trade-off.

Section~\ref{sec:4_7} analyzes supervision, grouping, and model relationships.

\subsection{Sensitivity to Supervision, Grouping, and Model Relationships}\label{sec:4_7}
For Q5, we vary the availability of task labels, the definition and number of excluded groups, and the relationship between collaborating models. These settings affect how much task supervision the target receives and what peer information can be reconstructed beyond the fitting groups.

\subsubsection{Supervision during Collaboration}

To separate the influence of labeled supervision from the inputs available for corroboration, we vary the labeled fraction while keeping the corroboration population fixed. Using 10\%, 25\%, 50\%, and 100\% of the training labels gives six-task mean gains of +1.54, +1.13, +0.81, and +0.52 percentage points over IA. With 50\%, 25\%, and 10\% of the collaboration labels, CL improves over the task-specific baselines selected from the strongest full-label results in Table~\ref{tab:full_comparison_means} by \(0.144 \pm 0.020\), \(0.326 \pm 0.024\), and \(0.633 \pm 0.030\) percentage points, respectively. At each label fraction, the task-mean gain is positive on all six tasks for every seed.

\subsubsection{Source-Grouping Strategies}

The excluded-group relation is intended to extend beyond the examples used to fit it, making group construction a natural factor to examine. On CrossRE \cite{ref40}, domain and size-matched random groups yield gains of +0.28 and +0.27 percentage points over IA. On Re-TACRED, predefined stratification increases the gain from +1.82 with random groups to +1.92. CL benefits from both strategies, with source-aware grouping providing further improvement.

\subsubsection{Model Relationships}

To examine whether collaboration requires architectural similarity, we compare the mixed CrossRE model pool with directed pairings involving structurally different ACE05 models. CrossRE gains +0.28 percentage points over IA across 12 directions. The heterogeneous ACE05 subset comprises the six directions between TagPrime and AMPERE, DEGREE, or PAIE and gains +0.03 on average. AMPERE and DEGREE each contribute approximately +0.18 when corroborating TagPrime. Gains therefore occur across model relationships, with architecture alone insufficient to determine useful complementarity.

\subsubsection{Corroboration partition}

The number of groups changes both the fitting population and the size of each held-out evaluation group. In the fixed-seed \(K\) sweep, CrossRE gains remain between +0.20 and +0.21 percentage points over IA for \(K=2,4,8,16,32\). Re-DocRED's gain increases by 0.14 percentage points as \(K\) rises from 2 to 32, while ACE05 achieves +0.13 at \(K=2\). Sensitivity to group granularity thus varies with task organization.

\subsection{Efficiency and Gain Analysis}\label{sec:4_8}

For Q6, we assess the cost of reconstructing cross-model responses during training and whether that cost carries into independent inference. Table~\ref{tab:efficiency} reports six-task mean training time, peak memory, and single-model inference latency relative to IA. CL uses \(2.23\times\) training time and \(1.04\times\) peak memory, with cross-model reconstruction confined to training. With training time matched to DML, CL retains a six-task mean gain of \(0.035 \pm 0.028\) percentage points over DML. Removing these components preserves the same independent-inference latency as IA with the matched response parameterization.

\begin{table}[htb]
\centering
\caption{Training and Inference Costs Across Six Tasks}\label{tab:efficiency}
\small
\begingroup
\setlength{\tabcolsep}{3pt}
\renewcommand{\arraystretch}{1.12}
\newcommand{\PFcost}[2]{\makecell[c]{#1\\[-1pt]{\scriptsize $(#2\times)$}}}
\begin{tabularx}{\linewidth}{@{}Xccc@{}}
\toprule
Method & \makecell[c]{Training time\\(min)} & \makecell[c]{Peak memory\\(GiB)} & \makecell[c]{Inference\\(ms/sample)} \\
\midrule
Independent Adaptation & \PFcost{251.1}{1.00} & \PFcost{25.0}{1.00} & \PFcost{17.5}{1.00} \\
DML & \PFcost{301.3}{1.20} & \PFcost{25.0}{1.00} & \PFcost{17.5}{1.00} \\
ADM & \PFcost{366.6}{1.46} & \PFcost{25.5}{1.02} & \PFcost{17.5}{1.00} \\
\textbf{Corroborative Learning} & \PFcost{560.0}{2.23} & \PFcost{26.0}{1.04} & \PFcost{17.5}{1.00} \\
\bottomrule
\end{tabularx}
\endgroup
\par\smallskip\noindent\parbox{\linewidth}{\footnotesize Main values are mean absolute costs across six tasks. Parentheses give ratios normalized to IA within each task before averaging. Training time is reported per hyperparameter configuration. Training time covers both targets of a model pair, with IA using the sum of two independent runs. Memory is the actual training peak. Measurements use an NVIDIA L40 in FP32. Inference is timed per independently deployed model with batch size 1.}
\end{table}

Finally, we relate the magnitude of the gains to the complementarity available before collaboration. On CrossRE, the low-complementarity group gains +1.18 percentage points over Checkpoint, versus approximately +1.36 for the other two groups combined. Together with the grouping and weight analyses, this links gains to available peer information and how it constrains target responses.

\subsection{Generalization to Visual Classification}\label{sec:4_9}

For Q7, we evaluate generalization from language tasks to visual classification using ResNet-50 \cite{ref75}, ConvNeXt-Tiny \cite{ref76}, and ViT-B/16 \cite{ref77}, forming three pairs and six directed combinations. Models are selected on a fixed 10k-image subset of the ImageNet-1K \cite{ref78} validation set and evaluated on the disjoint 40k subset and ImageNetV2 \cite{ref79}. The pretrained backbones remain fixed, and label fractions refer to the collaboration stage. CL achieves the highest mean Top-1 accuracy on both evaluation sets with either 100\% or 10\% collaboration labels, as reported in Table~\ref{tab:visual_classification}. With 10\% labels, it exceeds Diff-DML by 0.326 and 0.265 percentage points, respectively, with positive gains for every direction and seed.

\begin{table}[htb]
\centering
\caption{Visual Classification Accuracy at 100\% and 10\% Collaboration Labels}
\label{tab:visual_classification}
\small
\begingroup
\setlength{\tabcolsep}{1.5pt}
\renewcommand{\arraystretch}{1.12}
\newcommand{\VCscore}[2]{$#1_{\scriptstyle #2}$}
\begin{tabular*}{\columnwidth}{@{\extracolsep{\fill}}lcccc@{}}
\toprule
Method & \multicolumn{2}{c}{\makecell[c]{ImageNet-1K\\held-out 40k}} & \multicolumn{2}{c}{ImageNetV2} \\
\cmidrule(lr){2-3}\cmidrule(lr){4-5}
 & 100\% & 10\% & 100\% & 10\% \\
\midrule
Checkpoint & $81.460$ & $81.460$ & $70.907$ & $70.907$ \\
IA & \VCscore{81.663}{0.018} & \VCscore{81.527}{0.015} & \VCscore{71.039}{0.021} & \VCscore{70.948}{0.007} \\
KD & \VCscore{81.830}{0.013} & \VCscore{81.636}{0.009} & \VCscore{71.151}{0.016} & \VCscore{71.019}{0.007} \\
LogitStd & \VCscore{81.825}{0.025} & \VCscore{81.621}{0.021} & \VCscore{71.146}{0.033} & \VCscore{71.014}{0.011} \\
DIST & \VCscore{81.900}{0.038} & \VCscore{81.677}{0.034} & \VCscore{71.198}{0.044} & \VCscore{71.048}{0.023} \\
DKD & \VCscore{81.943}{0.014} & \VCscore{81.721}{0.001} & \VCscore{71.227}{0.023} & \VCscore{71.078}{0.015} \\
DML & \VCscore{81.913}{0.032} & \VCscore{81.731}{0.030} & \VCscore{71.209}{0.036} & \VCscore{71.087}{0.022} \\
MCL & \VCscore{81.872}{0.007} & \VCscore{81.686}{0.001} & \VCscore{71.182}{0.014} & \VCscore{71.058}{0.008} \\
ADM & \VCscore{81.947}{0.039} & \VCscore{81.728}{0.036} & \VCscore{71.232}{0.041} & \VCscore{71.084}{0.029} \\
Diff-DML & \VCscore{\underline{82.005}}{0.014} & \VCscore{\underline{81.757}}{0.009} & \VCscore{\underline{71.272}}{0.012} & \VCscore{\underline{71.103}}{0.009} \\
\midrule
\textbf{CL} & \VCscore{\mathbf{82.101}}{0.041} & \VCscore{\mathbf{82.084}}{0.033} & \VCscore{\mathbf{71.376}}{0.041} & \VCscore{\mathbf{71.368}}{0.025} \\
\bottomrule
\end{tabular*}
\endgroup
\par\smallskip\noindent\parbox{\linewidth}{\footnotesize Top-1 accuracy (\%). Results average six directions within each seed, then three collaboration seeds. Subscripts give the sample SD of the three seed means. IA and Checkpoint equally average the three targets. Checkpoint is fixed and has no seed SD. Bold and underlined means indicate the best and second-best results.}
\end{table}

With 10\% collaboration labels, CL outperforms tuned response regularization and self-reconstruction on both visual evaluation sets. Correct peer correspondence, source-excluded fitting, and reciprocal optimization each contribute to these gains.

With full collaboration labels, CL outperforms DML on both visual evaluation sets when their training times are matched.

\section{Conclusion}\label{sec:5}

Corroborative Learning uses peer representations to evaluate task-response changes across groups. Source-excluded reconstruction and joint optimization improve independently deployed models without directly matching outputs or features. The directional analysis explains how corroboration balances signal retention and noise suppression. Experiments on six language tasks with 24 models and 36 pairs, and on visual classification, connect gains to complementarity, corroboration quality, and supervision. Extending the current pairwise setting to automatic peer selection and larger model collections is a natural next step.

\FloatBarrier
\section*{Acknowledgments}
This work was supported by the National Natural Science Foundation of China under Grants 72231011 and 72001209, and by the Postdoctoral Fellowship Program of CPSF under Grants GZB20250994 and 2025M774445.

\end{document}